\documentclass{IEEEtran}
\usepackage{cite}
\usepackage{algorithm}
\usepackage{algpseudocode}
\usepackage{graphicx}
\usepackage{textcomp}
\usepackage{xcolor}
\usepackage{subfiles}
\usepackage{subcaption}
\usepackage{graphicx}
\usepackage{svg}
\usepackage{url}
\usepackage{hyperref}
\usepackage{comment}
\usepackage{amsmath, amssymb, amsfonts, amsthm, mathtools, dsfont}
\usepackage{algorithm,algpseudocode}
\usepackage{bm, multirow, array, xfrac}
\newtheorem{theorem}{Theorem}
\newtheorem{remark}{Remark}
\newtheorem{lemma}{Lemma}

\newtheorem{definition}{Definition}

\newtheorem{assumption}{Assumption}
\usepackage{color}
\newcommand\norm[1]{\lVert#1\rVert}

\def\BibTeX{{\rm B\kern-.05em{\sc i\kern-.025em b}\kern-.08em
    T\kern-.1667em\lower.7ex\hbox{E}\kern-.125emX}}
\begin{document}
\title{\LARGE \bf Energy-Constrained Resilient Multi-Robot Coverage Control}
\author{ 
Kartik A. Pant, Jaehyeok Kim, James M. Goppert, and Inseok Hwang
\thanks{The authors are with the School of Aeronautics and Astronautics, Purdue University,
West Lafayette, IN 47906. Email: (kpant, kim2153, jgoppert, ihwang)@purdue.edu}
}
\maketitle
\thispagestyle{empty}
\begin{abstract}
The problem of multi-robot coverage control becomes significantly challenging when multiple robots leave the mission space simultaneously to charge their batteries, disrupting the underlying network topology for communication and sensing. To address this, we propose a resilient network design and control approach that allows robots to achieve the desired coverage performance while satisfying energy constraints and maintaining network connectivity throughout the mission. We model the combined motion, energy, and network dynamics of the multirobot systems (MRS) as a hybrid system with three modes, i.e., coverage, return-to-base, and recharge, respectively. We show that ensuring the energy constraints can be transformed into designing appropriate guard conditions for mode transition between each of the three modes. Additionally, we present a systematic procedure to design, maintain, and reconfigure the underlying network topology using an energy-aware bearing rigid network design, enhancing the structural resilience of the MRS even when a subset of robots departs to charge their batteries. Finally, we validate our proposed method using numerical simulations.
\end{abstract}
\section{Introduction}
\label{sec:introduction}
The problem of multi-robot coverage control is to design a control law for a team of robots to cooperatively maximize the coverage of a given mission space. 
The existing methods either use gradient-based methods \cite{zhong2011distributed} or Voronoi partition-based methods \cite{cortes2004coverage} for coverage control problems. However, most of the existing work does not consider the energy constraints of the robots during the coverage task. In practice, since robots are battery-powered, they contain only a finite amount of energy for the mission. For example, most commercially available UAVs operate only for about $15-20$ minutes on a single charge of the batteries \cite{meng2018hybrid}. To efficiently cover a desired area under energy constraints, the robots must repeatedly cycle between the mission space and the recharging zone, which makes the control design challenging. Another challenge arising from robots' finite energy is the network disruption caused by robots' departure to recharge their batteries. Thus, designing a resilient coverage control strategy becomes essential to ensure robots don't deplete their batteries in the mission space and efficiently handle the network disruption caused by departing robots. 
\begin{figure}[t]
\centering
\includegraphics[width=0.43\textwidth,clip]{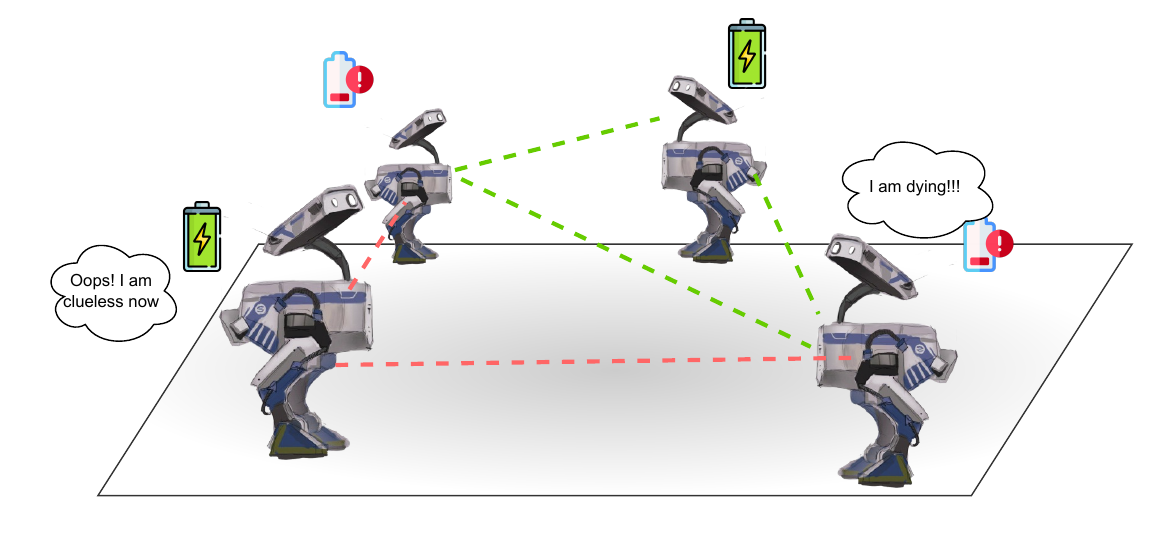}
\caption{Illustration of multi-robot coverage control with energy constraints. The underlying network gets disrupted when a subset of robots completely depletes their batteries.}
\label{fig:illustration}
\end{figure}

Some recent efforts have been made to design energy-aware control methods for multi-robot systems (MRSs) coordination \cite{setter2016energy}. The idea is to consider the energy dynamics of the system along with the motion dynamics. The authors in \cite{meng2018hybrid} utilize this idea for coverage control problems. In \cite{meng2018multi}, the authors proposed a framework to design optimal recharging levels in a coverage control setting. A comparison of centralized and decentralized coverage for energy-constrained systems is carried out in \cite{meng2021price}. However, the above studies have the following limitations: (i) they are limited to the unicycle dynamics, and (ii) they do not consider the network disruption caused by robot departure for battery recharging. 

In this paper, we present an energy-constrained resilient coverage control by combining the motion, energy, and network dynamics of the MRS. We model the combined energy and motion dynamics of the MRS as a hybrid system with switching modes. We split the robots' behavior into three different modes of operations: (i) coverage mode (energy depletion), (ii) to-charging mode, and (iii) recharging mode (energy repletion). To ensure that the robots do not run out of battery during the mission, we design guard conditions for mode transitions between coverage and to-charging modes. Additionally, we address the network reconfiguration problem that occurs when the robots leave for recharging. We propose an energy-aware network topology by leveraging the concepts from bearing rigidity theory. The idea is to construct a bearing-rigid network with well-distributed energy levels such that no two neighbours have the same energy levels. This allows the network to have the desired reconfigurability properties essential for resilient MRS operations. Finally, we utilize a distributed nonlinear MPC for the coverage task that simultaneously tracks Voronoi centroids \cite{carron2020model,cortes2004coverage} while maintaining the proposed bearing rigid network topology \cite{pant2025enhancing}.  
The remainder of this paper is organized as follows. Section \ref{sec:prelim} covers preliminaries on bearing rigidity. Section \ref{sec:problem_formulation} formulates the energy-constrained coverage problem. Section \ref{sec:method} presents the proposed resilient control architecture, including the hybrid model, and energy-aware network design and reconfiguration. Section \ref{sec:sim} shows numerical simulation results, and Section \ref{sec:conc} concludes the paper.

\textit{Notations:} 
Let $\mathcal{G} = \{ \mathcal{V}, \mathcal{E}\}$ denote an undirected graph with $\mathcal{V}$ as its vertices and $\mathcal{E} \subseteq \mathcal{V}\times\mathcal{V}$ as its edge set with $n = |\mathcal{V}|$ and $m = |\mathcal{E}|$. The set of neighbours of a vertex $i$ is denoted as $\mathcal{N}_i \triangleq \{j\in \mathcal{V}:(i,j)\in \mathcal{E} \}$.
\section{Preliminaries}
\label{sec:prelim}
In this section, we provide the necessary mathematical preliminaries from the bearing rigidity theory \cite{zhao2015bearing, trinh2019minimal}.
\subsection{Bearing Rigidity Theory}
Consider $n$ robots in $\mathbb{R}^d$ $(n\geq2, d\in\{2,3\})$ with their positions defined as $\{p_i\}_{i=1}^n$. The position vector $\mathbf{p} = [p_1^\top, \dots, p_n^\top]^\top \in \mathbb{R}^{dn}$ defines the MRS configuration. Given a configuration $\mathbf{p}$ along with an undirected graph $\mathcal{G}$ representing the underlying robot network, a framework is defined as $\mathcal{G}(\mathbf{p})$, where each vertex $i$ of the graph is uniquely mapped to the position $p_i$, respectively. Given a framework $\mathcal{G}(\mathbf{p})$, it is assumed that for each edge $e_{ij} \in \mathcal{E}$, both robots $i$ and $j$ can measure their respective bearings, where the bearing vector is $g_{ij}= \frac{p_{ij}}{\norm{p_{ij}}}$, where $p_{ij}:=p_j - p_i$ is the displacement vector.   
Define the bearing function $f_{B}: \mathbb{R}^{dn} \rightarrow \mathbb{R}^{dm}$ to describe all the bearings of the framework $\mathcal{G}(\mathbf{p})$ as: 
\begin{equation}
    \label{eq:bearing_func}
    f_B(\mathbf{p}) = [g_1^\top, \dots, g_m^\top]^\top,
\end{equation}
where $m=|\mathcal{E}|$ is the number of edges in the framework, $g_e$ denotes the bearing of edge $e \in \mathcal{E}$, and $\mathbf{p} = [p_1^\top, \dots, p_n^\top]^\top$.

Now, we present the definitions of infinitesimal bearing rigidity and generically bearing rigid graphs. 
\begin{definition}[Infinitesimal Bearing Rigidity (IBR) \cite{zhao2015bearing}]
An undirected framework $\mathcal{G}(\mathbf{p})$ formed by robots is infinitesimal bearing rigid (IBR) if its bearing motions are trivial, i.e., translation and scaling.
\end{definition}
\begin{definition}[Generically Bearing Rigid Graph \cite{trinh2019minimal}]
An undirected graph $\mathcal{G}$ is generically bearing rigid (GBR) if it allows at least one configuration $\mathbf{p}\in \mathbb{R}^{dn}$ that makes $\mathcal{G}(\mathbf{p})$ an IBR framework.
\end{definition} 

We now review the iterative procedure to construct GBR graphs using the Henneberg construction \cite{trinh2019minimal}. 
\begin{definition}[Henneberg Construction \cite{trinh2019minimal}]
Consider a graph $\mathcal{G} = (\mathcal{V}, \mathcal{E})$, a new graph $\mathcal{G^{'}} = (\mathcal{V}^{'}, \mathcal{E}^{'})$ is formed by adding a new vertex $v$ to the original graph by using any one of the following two operations:
\begin{enumerate}
    \item Vertex Addition: Pick two vertices $i,j\in\mathcal{V}$ from graph $\mathcal{G}$ and connect it to $v$. This will lead to the resulting graph $\mathcal{G}^{'}$ with $\mathcal{V}^{'}= \mathcal{V} \cup v$ and $\mathcal{E}^{'} = \mathcal{E}\cup\{(v, i),(v, j)\}$. 
    \item Edge Splitting: Pick three vertices $i,j, k\in\mathcal{V}$ and an edge $(i,j) \in \mathcal{E}$ from graph $\mathcal{G}$. Connect each vertex to $v$ and delete the edge $(i,j)$. In this case, the resulting graph $\mathcal{G}^{'}$ will have $\mathcal{V}^{'} = \mathcal{V} \cup v$ and $\mathcal{E}^{'} = \mathcal{E}\cup\{(v, i),(v, j), (v, k)\} \backslash \{ (i,j)\}$. 
\end{enumerate}
\end{definition}
\section{Problem Formulation}
\label{sec:problem_formulation}
We consider $n$ robots performing a coverage task in the mission space defined by a convex region $\mathcal{Q} \subset \mathbb{R}^d$, where $d$ is the dimension of the mission space.  Each robot's dynamics can be described as, 
\begin{align}
\label{eq:dyn}
    x_{i,k+1} &= f_i(x_{i,k}, u_{i,k}) \\
    p_{i,k} &=  C_{i}x_{i,k}
\end{align}
where $x_{i,k} \in \mathbb{R}^{n_x}$ is the state, $u_{i,k} \in \mathbb{R}^{n_u}$ is the control input, $p_{i,k} \in \mathbb{R}^d$ is the position, $f_i: \mathbb{R}^{n_x}\times\mathbb{R}^{n_u}\rightarrow \mathbb{R}^{n_x}$ denotes the dynamics of the $i^{th}$ robot, respectively. The matrix $C_i \in \mathbb{R}^{d \times n_x}$ picks the position from the state. The dynamics of the robots are assumed to be known and Lipschitz continuous. All robots are subject to state and input constraints, i.e., $(x_{i,k},u_{i,k})  \in \mathcal{Z}_i$ for all $k\geq 0$, where $\mathcal{Z}_i \subset \mathbb{R}^{n_{x}+n_{u}}$ is a compact set. 

Unlike most of the existing literature on multi-robot coverage control, we consider that the robots have a finite onboard energy supply, modeled by their battery's state-of-charge (SOC) denoted as $s_{i,k}$ (i.e., the fraction of the battery available at time $k$). We consider that for each robot, the energy consumption is a function of its current state and control inputs, e.g., the energy consumption for a quadrotor is proportional to the motor current which is proportional to the rotor speeds. This yields the following SOC dynamics:
\begin{equation}
\label{eq:energy_dyn}
    s_{i,k+1} = s_{i,k} - \mu \xi_i(x_{i,k}, u_{i,k}) - \gamma,
\end{equation}
where $\xi_i(\cdot)\colon\mathbb{R}^{n_x\times n_u} \rightarrow \mathbb{R}$ denote the energy consumption function, $\mu$ is a positive scaling constant to ensure that $s_{i,k}\in [0,1]$ and $\gamma$ is a fixed positive constant, denoting the constant energy degradation due to sensing and hovering. When $s_{i,k}\leq0$, robot $i$ is considered inactive in the mission space.
\begin{figure}[t]
\centering
\includegraphics[width=0.45\textwidth,clip]{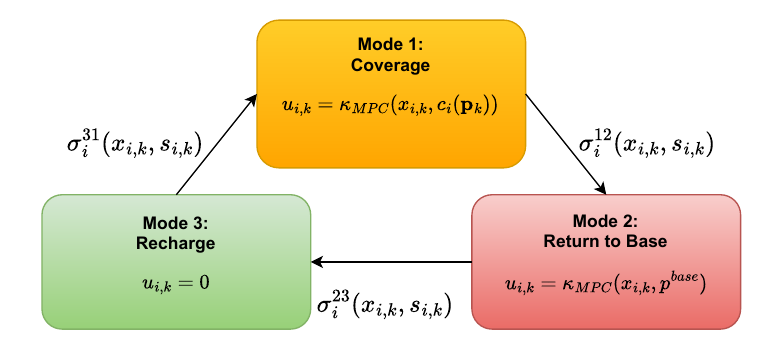}
\caption{Hybrid system model for coverage control with three modes: coverage, return-to-base, and recharge.}
\label{fig:hybrid}
\end{figure}

A charging station is available to replenish the robot's energy supply to prevent inactivity. The location of the charging station is denoted as $p^{base}$. The charging dynamics can be described as follows:
\begin{equation}
    s_{i,k+1} = s_{i,k} + \zeta_i(s_{i,k}),
\end{equation}
where $\zeta_i(\cdot)\colon\mathbb{R}\rightarrow\mathbb{R}$ is a charging function.
\begin{assumption}
\label{ass:recharge}
The recharging zone can handle multiple robots charging at the same time.    
\end{assumption}
In previous works \cite{setter2016energy, meng2018hybrid}, a constant charging rate is assumed. In practice, the charge rate decreases as the battery SOC increases \cite{bose2022study}. Thus, in this work, we consider a more general model for energy dynamics and charging. 
\begin{remark}
We consider a general model for the energy consumption function. More specific models for robots with the unicycle dynamics based on acceleration and velocities can be found in \cite{setter2016energy, meng2018hybrid}.  
\end{remark}

The problem is to design a distributed controller for the MRS, which can iteratively steer the robots to the centroidal Voronoi configuration for persistent coverage while satisfying the state, input, and energy constraints. 
Due to the energy dynamics, the problem transforms into a \emph{dynamic} multi-robot coverage scenario without any fixed equilibrium point, where the robots move back and forth between the coverage zone and the charging zone, unlike the existing work \cite{carron2020model}. Another challenge is to ensure that the remaining robots' network must be reconfigured appropriately for persistent coverage once a subset of the robots leave the mission space to recharge their batteries. 
To address these issues, in this work, we leverage concepts from hybrid systems modeling \cite{meng2018hybrid} and bearing rigidity theory \cite{zhao2015bearing, zhang2023self, trinh2019minimal} to construct a resilient control and network design strategy ensuring uninterrupted coverage operations while satisfying state, input, connectivity and energy constraints. Our control architecture ensures that multiple robots do not exhaust their batteries together, and the ones with critically low batteries reach the charging zone before exhausting all their batteries.

\section{Energy-Constrained Coverage Control}
\label{sec:method}
In this section, we describe the hybrid system modeling of the combined motion and energy dynamics of the MRS. We present the design of guard conditions for mode transitions between the \textit{coverage} and the \textit{return-to-base} modes, ensuring that the robots never run out of battery and reach the recharging station in the shortest time.
Then, we present a self-organized hierarchical network construction and reconfiguration procedure using bearing rigidity theory. This allows the robot network to repair its connectivity locally even when a subset of robots leave the network. Finally, we present a distributed tracking MPC control that enforces bearing maintenance (rigid network) and tracks the time-varying reference setpoints obtained using Voronoi partitioning, considering all active robots in the mission space. 
\subsection{Hybrid System Modeling}
Our main goal is to ensure that the energy of robots (SOC) stays positive at all times $k>0$, i.e., $s_{i,k}\geq0$. To ensure this, we first model the combined motion and energy dynamics as a hybrid system with three modes. All robots in an MRS are modeled into a directional hybrid system model, as shown in Fig. \ref{fig:hybrid}, where the robots cycle through in the following manner: Mode $1$ $\rightarrow$ Mode $2$ $\rightarrow$ Mode $3$ $\rightarrow$ Mode $1$, respectively. We refrain from invoking the complete hybrid automaton notations for the sake of notational simplicity. Each mode represents the combined motion and energy dynamics of the robot at different stages of the mission. 

In Mode $1$, the robot performs a coverage mission by steering itself towards the assigned waypoint determined by the Voronoi partitioning. Once the robot state hits the guard condition (switching condition) $\sigma_i^{12}(x_{i,k}, s_{i,k})$, the robots switch to Mode $2$, where it returns to the charging station to recharge its batteries. As soon as the robot arrive at the charging station, the guard condition $\sigma_i^{23}(x_{i,k}, s_{i,k})$ is triggered, and it switches to the charging mode. The robot remains in Mode $3$ until $s_i$ reaches the threshold $s_{th}$ and $\sigma_i^{31}(x_{i,k}, s_{i,k})$ is triggered. Lastly, the robot returns to Mode $1$ to return to the coverage mission. 
\subsection{Guard Conditions and Feasibility}
We now present the design of guard conditions for mode transitions between the coverage and to-charging modes. Before presenting the design of guard conditions, we first describe the following minimum-time open-loop trajectory optimization problem. This will enable us to design guard conditions that ensure the robots always satisfy energy constraints and reach the recharging station in the shortest time possible. The idea is to compute the energy summation along the minimum-time reference trajectory to design the guard conditions that ensure the energy feasibility and guarantee the robot reaches the recharging station in the minimum time.  
\begin{subequations}
\label{eq:min_traj}
\begin{align}
    \min_{(\mathbf{\bar{x}}_i, \mathbf{\bar{u}}_i, \tau_i)} \quad  &\tau_i\\
    &\text{s.t.} \quad C_ix_{i,\tau_i} \in p^{base}, \quad l = \{1,\dots, \tau_i\},\\
    &x_{i,l+1} = f_i(x_{i,l}, u_{i,l})\\
    &(x_{i,l}, u_{i,l}) \in \mathcal{Z}_i\\
    &d(x_{i,l},x_{j,l})\leq 0, \ \forall j \in \{1, \dots, n\}, \ j \neq i.
\end{align}
\end{subequations}
where $d(\cdot)$ is the safety constraint among the robots in the mission space. Let the solution of the above optimization problem be denoted as $(\mathbf{\bar{x}}_i^*, \mathbf{\bar{u}}_i^*, \tau_i^*)$. We now present our main result on designing guard conditions, as stated in Theorem \ref{th:guard_cdn}, which ensures that robots have sufficient energy to reach the charging zone before their energy levels are fully exhausted.
\begin{theorem}
\label{th:guard_cdn}
Suppose each robot in an MRS satisfying the dynamics \eqref{eq:dyn} at time $k$, with the recharging station satisfying Assumption \ref{ass:recharge}, has the following guard condition to switch from Mode $1$ to Mode $2$.
\begin{equation}
\label{eq:guard_cdn}
    \sigma_i^{12}(s_{i,k}, x_{i,k}) \coloneq s_{i,k} - \underbrace{\Biggl(\sum_{l=k}^{k+\tau_i^*}  \mu \xi_i(\bar{x}^*_{l,k}, \bar{u}^*_{l,k}) + \gamma\Biggr)}_{s_{i,k}^*}
\end{equation}
Then, each robot's energy dynamics will always satisfy $s_{i,k}\geq0$ throughout the coverage mission. 
\end{theorem}
\begin{proof}
It is clear from the hybrid system modeling that the robots can exhaust their batteries, i.e., $s_{i,k}\leq0$ only in Modes $1$ and $2$. Thus, if we can ensure that the transition from Mode $1$ to Mode $2$ leaves sufficient battery in each robot to reach the recharging point, we can guarantee that the robots will never exhaust their batteries in the mission space. 
Using \eqref{eq:energy_dyn}, we observe that the discharge energy dynamics of the robot is given as,
\begin{equation}
\label{eq:energy_dyn_thm}
    s_{i,k+1} = s_{i,k} - \mu \xi_i(x_{i,k}, u_{i,k}) - \gamma,
\end{equation}
Summing \eqref{eq:energy_dyn_thm} along the minimum time trajectory and assigning $s_{i,k+1} = 0$, we get the desired guard condition \eqref{eq:guard_cdn}.

Now, we observe that the robots in Mode $1$ will have $s_{i,k}\geq s_{i,k}^*$ and thus, $s_{i,k}\geq0$ is satisfied in Mode $1$. In Mode $2$, as a consequence of the guard condition, the robots will have sufficient energy to follow the optimal minimum time trajectory to the recharging station. Hence, $s_{i,k}\geq0$ will be satisfied and the robots will reach the recharging zone in the shortest time. 
\end{proof}
\begin{remark}
While our guard condition is designed based on the optimal minimum-time trajectory to ensure energy constraint feasibility and shortest arrival time to the charging station, alternative objectives, i.e., minimizing energy consumption, can also be considered.
\end{remark}
\begin{remark}
We note that our proposed method may be challenging to implement for a resource-constrained MRS. However, it provides theoretical foundations for systematically addressing energy constraints in MRS coverage problems.
\end{remark}
The design of the guard condition $\sigma_i^{23}(s_{i,k}, x_{i,k})$ is decided based on the scheduling at the recharging station. In this work, for the sake of simplicity, we assume that the recharging station can handle multiple robots simultaneously. Thus, the guard condition is triggered when the robots reach the recharging zone. The guard condition $\sigma_i^{31}(s_{i,k}, x_{i,k})$ is triggered only when the robots are charged to a constant maximum energy level. In general, the design of the guard conditions $\sigma_i^{23}(s_{i,k}, x_{i,k})$ and $\sigma_i^{31}(s_{i,k}, x_{i,k})$ can be arbitrary complex based on mission constraints, robots, and the facility in the charging zone. We defer those complications to our future work.  
\subsection{Energy-Aware Hierarchical Bearing-Rigid Network Design}
Without loss of generality, the network topology for communication and inter-robot sensing is assumed to be the same. We take inspiration from Henneberg's construction to iteratively construct the minimally bearing rigid graph for the underlying network design. However, the key difference lies in the choice of neighboring robots for each robot. We restrict robots with the same energy levels from clustering together to ensure that they do not leave the network simultaneously. The main idea is to evenly distribute the robots with low energy levels in the network, thereby minimizing the effect of the departure of multiple robots on the resulting network. 
\begin{figure}[t]
\centering
\includegraphics[width=0.36\textwidth,clip]{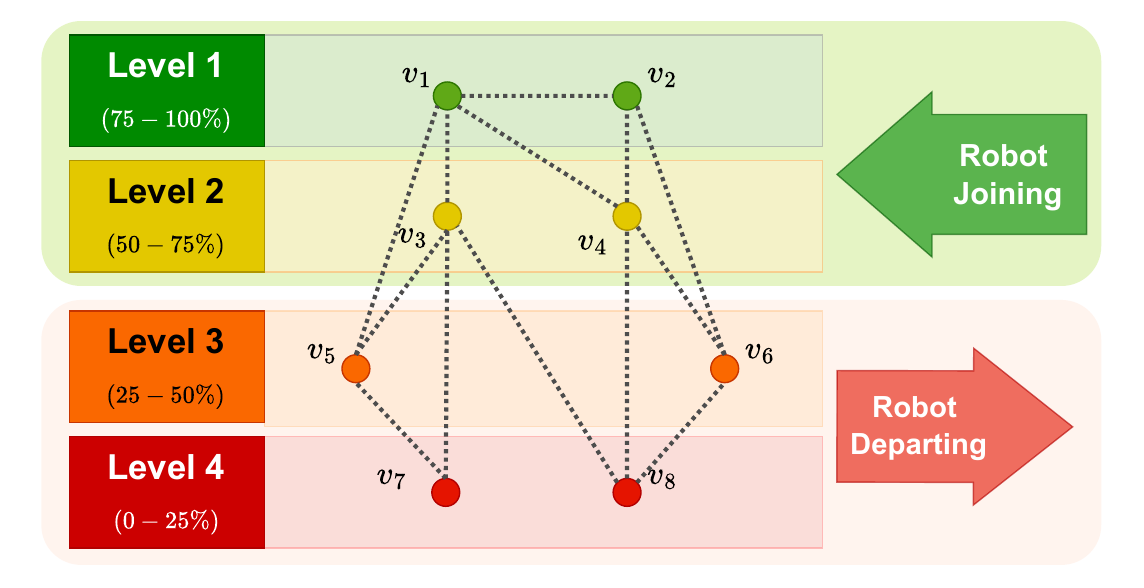}
\caption{Self-organized hierarchical bearing-rigid network using energy levels as hierarchy.}
\label{fig:network}
\end{figure}
\begin{algorithm}[t]
\caption{Construction of self-organized hierarchical energy-aware bearing-rigid graph and framework}\label{alg:construction}
\begin{algorithmic}[1]
\Require Robot positions $\mathbf{p}_k$ and their SOC $\mathbf{s}_k$ at time $k$.\\
$i\leftarrow0$; \\
Choose two robots from level $1$ as vertex $v_1$ and $v_2$;\\
$\mathcal{G}_k \leftarrow$ add vertex $v_1$ and $v_2$, and edge $e_{1,2} = (v_1, v_2)$;\\
$i\leftarrow i+2$;
\For {$i =2\dots,n$}
    \State Denote the $i^{th}$ robot as vertex $v_i$;
    \If{Vertex addition is used}
        \State Choose two robots from $\mathcal{G}_k$ as vertex $v_j$ and $v_k$ \\
        \qquad \quad such that $\eta(s_{v_j, k}) \leq \eta(s_{v_i, k})\leq \eta(s_{v_k, k})$;
        \State $\mathcal{G}_k \leftarrow$ add vertex $v_i$, and edges $e_{ij}$ and $e_{ik}$;
    \ElsIf{Edge splitting is used} 
        \State Choose a robot $v_j$ from $\mathcal{G}_k$ with at least two \\\qquad \quad neighbors $v_k$ and $v_p$ such that $\eta(s_{v_j,k}) \leq \eta(s_{v_i,k})$;
        \State Choose a robot $v_l$ from $\mathcal{G}_k$ such that\\\qquad \quad $\eta(s_{v_i,k}) \leq \eta(s_{v_l,k})$;
        \State $\mathcal{G}_k \leftarrow$ remove $e_{kj}$, add vertex $v_i$, and edges $e_{ij}$,\\
        \qquad \quad $e_{ik}$ and $e_{il}$;       
    \EndIf
\EndFor \\
\Return $\mathcal{G}_k$, $\mathcal{G}_k(\mathbf{p}_k)$;
\end{algorithmic}
\end{algorithm}

The idea is to first divide the robots into an energy hierarchy of four levels, i.e., level $1$ robots with $0-25\%$ battery, level $2$ robots with $25-50\%$ battery, and so on, as shown in Fig. \ref{fig:network}. Define $\eta(\cdot): [0,1] \rightarrow \{1,2,3,4\}$ as the energy level function, which takes in $s_{i,k}$ as input and provides the energy level as output. Then, when making new connections, it is ensured that each robot $i$ with energy level $\eta(s_{i,k})$ has at least two neighbors $u$ and $v$, one from one level up and the other from one level down, i.e., $\eta(s_{u,k}) \leq \eta(s_{i,k}) \leq \eta(s_{v,k})$. The resulting design procedure ensures that neighboring robots do not leave the network at the same time. Since the network design procedure is sequential, the same logic can be applied to robots joining the network after recharging their batteries. The overall construction of the network topology is provided in Algorithm \ref{alg:construction}.    
\subsection{Nonlinear Tracking MPC with Bearing Maintenance}
One of the underlying requirements for the network design presented earlier is to ensure that the robots maintain a constant bearing with their neighbors during the coverage task. To enforce the bearing constraint, we utilize the nonlinear tracking MPC framework to track the centroids while maintaining a constant bearing (see \cite{pant2025enhancing} for more details). Define the computed reference set point as $r_{i,k} \coloneq c_i (\mathbf{p}_{k})$, where $\mathbf{p}_k$ denotes the robots' position at time $k$ and $c_i (\mathbf{p}_{k})$ denotes the Voronoi centroid.

To account for bearing maintenance, we add bearing maintenance costs along with the Voronoi centroid tracking cost in the overall optimization process. We leverage the computed Voronoi centroids, denoted by $r_k$ and the underlying graph $\mathcal{G}_k$ at time $k$, to establish a bearing-rigid formation. This helps us in determining the bearing vector for each of the active robots in the MRS at time $k$. More specifically, 
given the positions $\mathbf{p}_k$ and the SOCs $\mathbf{s_k}$ of the active robots, we obtain a self-organized hierarchical network topology $\mathcal{G}_k$ using Algorithm \ref{alg:construction}. The desired bearing vector for robot $i$ at time $k$, denoted as $g_{i,k}$, is computed using the bearing function $g_{i,k} = f_B(\mathbf{r}_k)$. Our objective is to design a control law that steers the robots to the desired time-varying setpoints $r_{i,k}$ while maintaining a constant inter-robot bearing vector equal to $g_{i,k}$.

Define $J_i(x_{i,k}, r_{i,k}, g_{i,k}; \mathbf{u}_i, \bar{r}_i)$ as the MPC cost function, where $x_{i,k}, \ r_{i,k}$, and $g_{i,k}$ are inputs, and $\mathbf{u}_i$ and $\bar{r}_i$ are outputs. For simplicity, we define $J_i(\cdot) \coloneq J_i(x_{i,k}, r_{i,k}, g_{i,k}; \mathbf{u}_i, \bar{r}_i)$.
Let $N$ be the prediction horizon, and $x_{i,l|k}$ and $u_{i,l|k}$ as the $l$-step forward prediction of the state and input of robot $i$ at time $k$, respectively, with $l \in \{1,\dots, N\}$, where $x_{i,0|k} = x_{i,k}$ and $\mathbf{u}_i = [u_{i,0|k}^\top, u_{i,1|k}^\top, \dots, u_{i,N|k}^\top]^\top$. The overall cost function comprising the tracking cost and the bearing maintenance cost is defined as:
\begin{align}
\label{eq:mpc_cost}
 J_i(\cdot) = \sum_{l=0}^{N-1} &\ell_{i,l}(x_{i,l|k} - \bar{x}_{i}, u_{i,l|k} -  \bar{u}_i) + \ell_{i,N}(x_{i,l|k} - \bar{x}_i, \bar{r}_i) \nonumber
 \\+ \mu &\ell_{i,r}(r_{i,k} - \bar{r}_i) + (1-\mu)\ell_{i,b}(g_{i,k}, \bar{r}_i),    
\end{align}
where the stage cost and the terminal cost are $\ell_{i,l}: \mathbb{R}^{n_x} \times \mathbb{R}^{n_u} \rightarrow \mathbb{R}$ and $\ell_{i,N}: \mathbb{R}^{n_x} \times \mathbb{R}^{d} \rightarrow \mathbb{R}$, respectively. Furthermore, centroid tracking costs and bearing maintenance costs are defined as $\ell_{i,r}: \mathbb{R}^{d} \rightarrow \mathbb{R}$ and $\ell_{i,b}: \mathbb{R}^{dn_x} \times \mathbb{R}^{d} \rightarrow \mathbb{R}$, respectively. All of the above-mentioned cost functions are positive-definite. $\mu\in(0,1]$ denotes a constant to ensure that the combined reference tracking and bearing maintenance costs are convex. It is important to note that the first two terms of $J_i(\cdot)$ penalize the deviation of the state from the artificial position reference $\bar{r}_i$, which is also a decision variable. Whereas, $\ell_{i,r}(\cdot)$ penalizes the deviation between the actual reference centroid and the artificial reference $(r_{i,k} - \bar{r}_i)$. The overall nonlinear tracking MPC problem formulation can be described as:
\begin{subequations}
\label{eq:mpc}
\begin{align}
    \min_{u_i, \bar{x}_i, \bar{u}_i, \bar{r}_i} &J_i(x_{i,k}, r_{i,k}, g_{i,k}; \mathbf{u}_i, \bar{r}_i)\\
    &\text{s.t.} \quad l = \{1,\dots, N-1\}\\
    &x_{i,0|k} = x_{i,k}\\
    &x_{i,l+1|k} = f_i(x_{i,l|k}, u_{i,l|k})\\
    &(x_{i,l|k}, u_{i,l|k}) \in \mathcal{Z}_i\\
    &(x_{i,N|k}, \bar{r}_{i}) \in \Gamma_i\\
    &\bar{r}_i = C_i\bar{x}_i\\
    &\bar{x}_i = f_i(\bar{x}_i, \bar{u}_i),
\end{align}
\end{subequations}
where $\Gamma_i$ denotes the invariant terminal set for centroid tracking.
Each robot then independently determines its control input using a receding-horizon approach. Specifically, at time $k$, the distributed MPC control law for robot $i$ is defined as $ \kappa_{MPC}(x_{i,k}, r_{i,k}, g_{i,k}) = u_{i,0|k}^{*} \ \forall i \in \{1,\dots, n\}$, where $u^{*}$ is the optimal solution of \eqref{eq:mpc}. The robots apply the first control inputs from the optimal inputs resulting from solving the optimal control problem.
\section{Network Reconfiguration During Recharging}
\label{sec:reconfig}
\begin{algorithm}[t]
\caption{Reconfiguration of network after robot departure}\label{alg:recovery}
\begin{algorithmic}[1]
\Require Number of departing robots $n_d$, bearing rigid network framework $\mathcal{G}_k(\mathbf{p_k})$.
\State Divide the departing robots into levels $3$ and $4$.
\For {$i =1,2\dots,n_d$}
    \If{\texttt{energy\_level}== $4$}
        \State Use the reverse of the robot addition in Hennberg's \\
        \qquad \quad sequence. 
    \Else
        \State Find a contractible edge using Lemma \ref{lem:non_contract} in the  \\
        \qquad \quad level $1$, if none exist, then find it in level $2$ to add \\
        \qquad \quad new edges for all neighbors of the departing robot.     
    \EndIf
\EndFor
\end{algorithmic}
\end{algorithm}
In this section, we present the network reconfiguration mechanism when a subset of robots leaves the network to recharge their batteries. As a consequence of the energy-aware network design presented in the previous section, the resulting network topology possesses a unique local reconfiguration capability, where robots require connecting to their $2-$hop neighbors to regain the rigidity property. Furthermore, as an energy-aware hierarchy is established and maintained throughout the coverage task, we can leverage the inheritance property to repair the rigidity. This enables rigidity recovery even when multiple robots leave the network to recharge their batteries. Note that by construction, the network design ensures that the robot departure only occurs at level $3$ and level $4$, robots with energy levels less than $50\%$ of their total energy. We use the principle of edge contraction \cite{fidan2010closing, pant2025enhancing} to identify the set of new edges for the rigidity recovery of the framework. In edge contraction, two adjacent vertices $v_1$, $v_2 \in \mathcal{V}$ of a graph $\mathcal{G} = (\mathcal{V}, \mathcal{E})$ are merged into a single vertex by ``contracting" the edge between them. The new vertex should be adjacent to all the neighbors of $v_1$ and $v_2$ without redundant edges (if any exist). An edge is called a contractible edge if the resulting graph after the edge contraction operation is rigid. The following lemma provides sufficient conditions for identifying a contractible edge in the network. 
\begin{lemma}[Edge Contraction\cite{fidan2010closing}]
\label{lem:non_contract}
Consider a rigid graph $\mathcal{G} = (\mathcal{V}, \mathcal{E})$ in $d=2$. Let $v,w \in \mathcal{V}$ and $(v,w) \in \mathcal{E}$. The edge $(v,w)$ is non-contractible in $\mathcal{G}$ if multiple vertices (more than one vertex) are adjacent to both $v$ and $w$.
\end{lemma}
\begin{figure*}[t]
\centering
        \centering
        \includegraphics[height = 0.225\textwidth, width=0.7\textwidth, trim={1.4cm 0.4cm 1.5cm 0.2cm},clip]{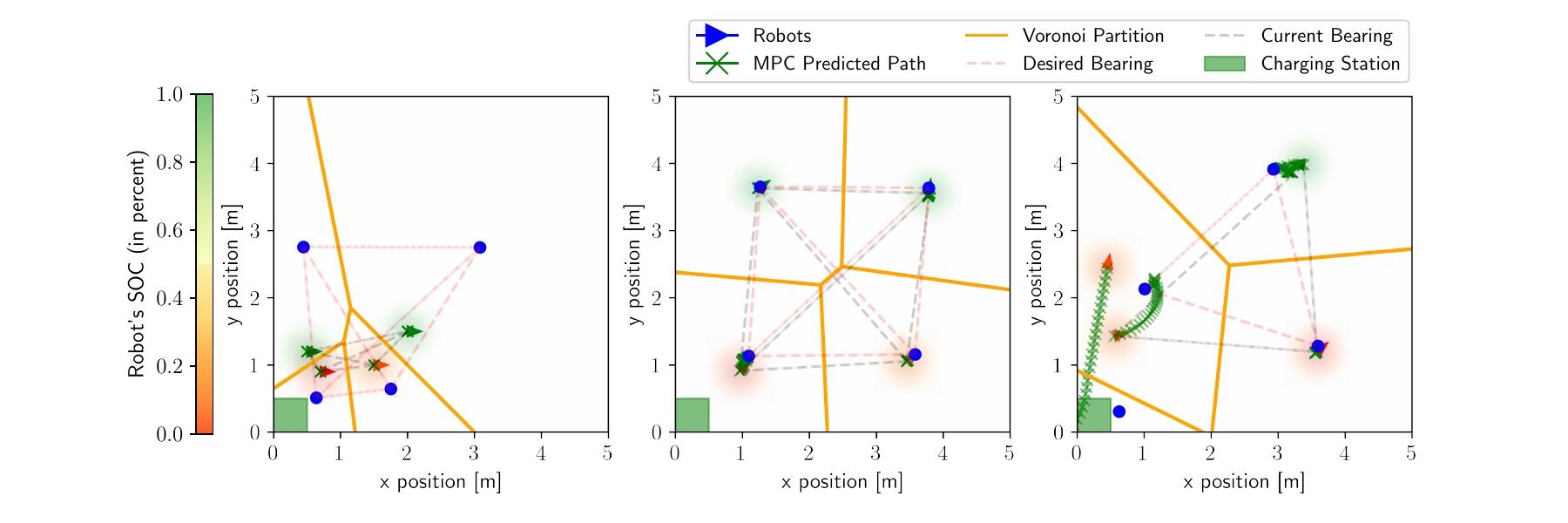}
        \includegraphics[height = 0.205\textwidth, width=0.28\textwidth,trim={0.2cm 0.4cm 0.3cm 0.2cm},clip]{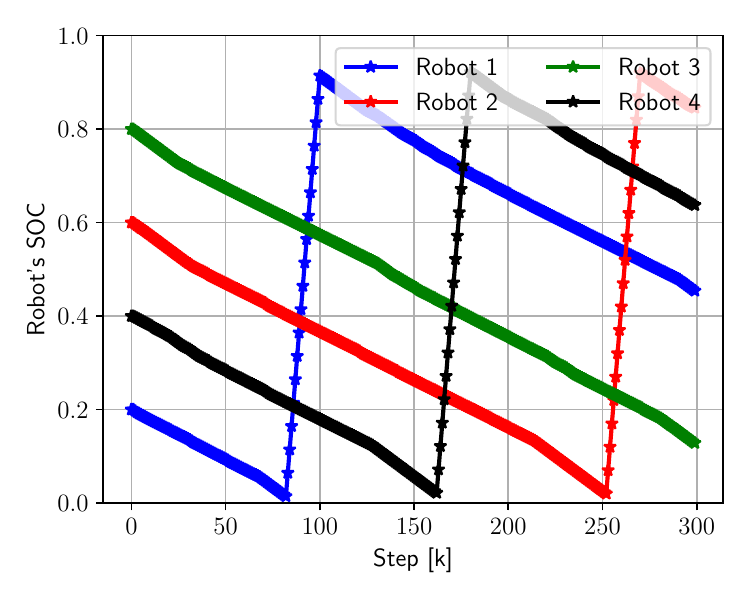}
\caption{\textbf{Left:} Coverage control with network reconfiguration when a robot departs to charge its batteries. \textbf{Right:} Evolution of robot energy levels (SOC) with time.}
\label{fig:coverage_energy}
\end{figure*}
The reconfiguration procedure is performed in a bottom-up fashion. This allows us to treat multiple robot departures as a sequence of single robot departures from levels $3$ and $4$. We now present the reconfiguration procedure for robots of these two levels separately as follows:

\textbf{Level 4 Robot Departure:} The rigidity recovery procedure is easier for robots at level $4$ compared to robots at level $3$. The idea is to invoke the reverse of the robot addition operation in the Henneberg sequence. Thus, if the departing robot has only two edges, those edges are removed from the graph $\mathcal{G}_k$. If the robot has three edges, then the reverse edge splitting operation is performed on the neighbors such that three edges are removed and an edge is added.   

\textbf{Level 3 Robot Departure:} Once all robots at level $4$ are removed, we shift our focus to level $3$ robots. The core idea for reconfiguring the network is to utilize the inheritance property due to the inherent hierarchy preserved by the underlying network. As all robots in level $3$ are restricted to having at least one neighbor in level $1$ or level $2$, as soon as a level $3$ robot leaves the network, its active neighbors in level $3$ and level $4$ inherit its level $1$ and level $2$ neighbors. 

The overall reconfiguration procedure is presented in Algorithm \ref{alg:recovery}.

\section{Numerical Simulations}
\label{sec:sim}
In this section, we show the effectiveness of our proposed method in maintaining a positive SOC during the entire coverage mission and swift reconfiguration of robots during recharging using numerical simulations. The coverage environment is defined as $\mathcal{Q} = [0,5] \times [0,5]$. The density function is chosen as $\phi(q) = 1$. We consider $4$ robots, each modeled as the unicycle dynamics. The dynamics can be represented as,
\begin{align}
    p^x_{i,k+1} &= p^x_{i,k} + T_s \cos (\theta_{i,k}) v_{i,k}\\
    p^y_{i,k+1} &= p^y_{i,k} + T_s \sin (\theta_{i,k}) v_{i,k}\\
    \theta_{i,k+1} &= \theta_{i,k} + T_s \omega_{i,k},
\end{align}
where $p_{i,k}^x$ and $p_{i,k}^y$ denote the positions and $\theta_{i,k}$ denotes the yaw of the $i^{\text{th}}$ robot at time $k$, respectively. Let $x_{i,k} = [p_{i,k}^x, p_{i,k}^y, \theta_{i,k}]^\top$denote the state of robot $i$. The control inputs are the linear and angular velocities denoted by $v_{i,k}$ and $\omega_{i,k}$ for each robot, respectively. The sampling time is $T_s = 0.1s$. The matrix $C_i$ is given by $
C_i = \begin{bsmallmatrix}
    1 & 0 & 0 & 0\\
    0 & 1 & 0 & 0
\end{bsmallmatrix}$. 
The position of each robot is constrained to be within the coverage region $\mathcal{Q}$ and the yaw $\theta_{i,k} \in [-\pi, \pi]$. The linear and angular velocities for each robot are limited to $\norm{v_{i}}\leq 1$ $m/s$ and $\norm{\omega_{i}}\leq \pi/2$ $rad/s$, respectively. The initial positions of all the robots are $p_{1,0} = [0.7,0.9]^\top$, $p_{2,0} = [0.5,1.2]^\top$, $p_{3,0} = [2.0,1.5]^\top$, and $p_{4,0} = [1.5, 1.0]^\top$. Additionally, we have the initial SOC of the robots as  $s_{1,0} = 0.2$, $s_{2,0} = 0.6$, $s_{3,0} =0.8$, and $s_{4,0} = 0.4$. The recharging zone is located at $p^{base} = [0.0, 0.0]^\top$. The stage cost is given by $\ell_i(x_i-\bar{x}_i,u_i - \bar{u}_i) = \norm{x_i - \bar{x}_i}^2_{Q_i} + \norm{u_i - \bar{u}_i}^2_{R_i}$, where $Q_i = \text{diag}(1,1,1)$ and $R_i= \text{diag}(0.05, 0.005)$ are chosen as diagonal matrices. The reference tracking cost and bearing maintenance costs are given by $\ell_{i,r}(r_i-\bar{r}_i) = \norm{r_i - \bar{r}_i}^2_{Q^r_i}$ and $\ell_{i,b}(g_i, \bar{r}_i) = \sum_j^{\mathcal{N}_i} \norm{g_{ij} - f_B(\bar{r}_i)}^2_{Q^b_{ij}}$, respectively, where $Q^r_i = I$ and $Q^b_{ij} = 100\cdot I$. The guard condition between coverage and to-charge modes is designed using minimum time straight-line trajectories to the recharge zone.

Figure \ref{fig:coverage_energy} shows the evolution of the robots for the first $90$ steps of a $300$ steps simulation, with the initial configuration at $k=0$, the intermediate configuration at $k=50$, and the first reconfiguration of the network at $k=90$. The right side in Fig. \ref{fig:coverage_energy} depicts the evolution of the robot's energy over time. We can observe that the robots never exhaust their batteries in the mission space as a result of our designed guard conditions.  

\section{Conclusion}
\label{sec:conc}
In this paper, we introduced a resilient coverage control algorithm for energy-constrained MRSs. We leveraged a hybrid system modeling to model the motion and energy dynamics of the MRS and transformed the robots' energy constraints to design suitable guard conditions. We developed an energy-aware hierarchical network design with local repair to ensure network rigidity after robots leave for charging. We presented a distributed MPC control law to maintain the network topology and perform the coverage task. Finally, we presented the numerical simulations to corroborate our theoretical results.

\bibliographystyle{ieeetr}
\bibliography{IEEEabrv,refs} 
\end{document}